\documentclass{IEEEcsmag}

\usepackage[colorlinks,urlcolor=blue,linkcolor=blue,citecolor=blue]{hyperref}

\usepackage{upmath}

\usepackage{color,array}
\usepackage{graphicx}
\usepackage{cite}
\usepackage{multirow}
\usepackage{amsmath}
\usepackage{amssymb}
\usepackage{booktabs}
\usepackage{makecell}

\jvol{XX}
\jnum{XX}
\paper{8}
\jmonth{March}
\jname{IT Professional}
\pubyear{2022}

\begin{document}

\title{Water and Sediment Analyse Using Predictive Models}

\author{Xiaoting Xu}
\affil{School of Computer Science, The University of Sydney}

\author{Tin Lai}
\affil{School of Computer Science, The University of Sydney}

\author{Sayka Jahan}
\affil{Department of Environmental sciences, Macquarie University}

\author{Farnaz Farid}
\affil{School of Social Sciences, Western Sydney University}

\markboth{Department Head}{Paper title}

\begin{abstract}
    The increasing prevalence of marine pollution during the past few decades motivated recent research to help ease the situation. Typical water quality assessment requires continuous monitoring of water and sediments at remote locations with labour intensive laboratory tests to determine the degree of pollution. We propose an automated framework where we formalise a predictive model using Machine Learning to infer the water quality and level of pollution using collected water and sediments samples. One commonly encountered difficulty performing statistical analysis with water and sediment is the limited amount of data samples and incomplete dataset due to the sparsity of sample collection location. To this end, we performed extensive investigation on various data imputation methods' performance in water and sediment datasets with various data missing rates. Empirically, we show that our best model archives an accuracy of 75\% after accounting for 57\%  of missing data. Experimentally, we show that our model would assist in assessing water quality screening based on possibly incomplete real-world data.
\end{abstract}

\maketitle

\section{Introduction}

\chapterinitial{Sediments}
are natural particles that develop as earth materials are broken down through weathering and erosion. Metal concentration is the standard indicator in marine water and sediments that denotes the sign of pollution. Due to the rapid development of industry and global urbanization, the pollution problem has raised mass attention worldwide. 
The impact of metal on the water and sediment quality is remarkably negative. Metals in water and sediments and hefty metals are persistent sources that may cause various adverse outcomes to creatures on the earth. Containing heavy metals in marine water sediments may result in transcriptional effects on stress-responsive genes\cite{1}. Heavy metal pollution will affect a few species and the whole ecosystem and every species in the ecosystem. There have been many studies about the flow of water and sediment from various aspects, for example, continuously monitors of heavy metal variability in highly polluted rivers~\cite{2}. 

Accumulation and prediction of metals in water and sediments are a complicated issue, where we propose to perform an empirical investigation on a real-world dataset by utilising predictive Machine Learning (ML) models. All follow-up actions of environmental policy and related restrictions can only be formulated after a clear assessment of water and sediment quality. Therefore, it is essential to evaluate the predictive capability of our model on water and sediment quality. 

In this paper, we performed an empirical evaluation of the water and sediment dataset collected in Australian ports located in six different areas \cite{3}. 
It is practically impossible to monitor all metals and make regulations  \cite{4}. Using an Artificial Intelligent learning-based approach helps to identify the essential pollutant in a data-driven approach, which helps to assist the pollution regulation. We tested the water and sediment samples and identified the primary pollutants. Several pollution indexes were used to determine the water and sediment quality and degree of pollution. We empirically combined different water and sediments sources to learn the correlation between various sediment content levels and their contribution to water pollution. Experimentally, we show that our model can achieve state-of-the-art predictive capability to identify highly polluted ports based purely on collected data.

\section{Related Works}

Heavy metal pollution has brought an unpredictable threat to aquatic ecosystems as an increasing population and industrialization expanded. With the rapid change in industrialization, the concentration of heavy metals is serious in sewage wastewater, industrial wastewater discharges and atmospheric deposition\cite{1},\cite{5}. Precisely, heavy metal concentrations in soil and water and sediments are becoming severe due to intensive human activities\cite{6}. According to Sather Noor Us’ research on the comparison between heavy metal elements, excessive levels of heavy metals (such as Fe, Cu, Zn, Co, Mn, Se and Ni) tend to be harmful to marine or marine life; other metals (such as Ag, Hg and Pb) are fatal to marine organisms\cite{7}.

Traditional approaches, such as geochemical methods like Inductively coupled plasma mass spectrometry (ICP-MS) requires a labour-intensive process where it is time-consuming and high in cost \cite{5}, \cite{8}. Moreover, these methods are not suitable when the test scale is substantially significant \cite{9}. 
In search of the ability to detect contamination in different areas, one possible approach is to combine multiple sources of contamination datasets in a meaningful way. Utilising multiple sources of information can enhance Machine Learning models' predictive capability and tackle the typical data scarcity issue with water sediment datasets.

Machine learning models can explore correlations between various variables more effectively and thus make more accurate predictions. For example, an artificial neural network (ANN) can classify images or recognize speech when conducting biological research \cite{10}. When there exists a mismatch of data features between datasets, data-imputation can tackle the problem of missing data
\cite{11}. Environmental indexes are some of the established standards to deduce the degree of water contamination\cite{3}. However, there exist multiple standards, such as geo-accumulation index (Igeo), enrichment factor (EF), and none are considered as the "golden standard". Therefore, the combination of dataset and environmental indexes is a possible approach to combine the well-established environmental index with the predictive capability of Machine Learning.

\section{Motivation}

The focus of this study is to train several machine learning and deep learning models, using water and sediment samples data pre-trained with numerous public sediments datasets. We focus on assessing marine water and sediment quality in several locations with varying water depths by extracting features from the ecological state of these areas. The objectives include (i) collecting related sediment datasets at vastly separated places to widen the sample size; (ii) assessing the most suitable missing data imputation method which is typical in marine datasets; and (iii) develop a set of machine learning and deep learning predictive models for classifying the water quality. 

In this research, we focus on two major challenges. The first is determining the extent of water pollution by using pollution indices derived from metal sediments. Current studies centred around explaining and calculating pollution indices but does not focus on the relationship between pollution indices and water and sediment quality. Our proposed model would tackle this issue by extracting the correlation between these indices to determine water and sediment quality.
The second lies in identifying a robust set of data imputation methods that can perform consistently across multiple sets of marine water datasets. Sediment contents tend to require specific types of equipment for detection, which means studies conducted by different groups might not be complete. This difficulty can be mitigated by utilising missing data imputation methods.

\textbf{Research Scope}
The scope of this research involves assessing heavy metals pollution in the sediments of marine water. In the broad sense, this topic covers many fields such as environmental science, biological science, oceanic science, and data science. We focus on the scope of data science. We utilize publically available datasets to train a predictive model, then perform an extensive feasibility study on the predictive performance on an unseen private dataset.  The expected outcome is to develop a machine learning model that uses heavy metals indices as input attributes and automatically outputs the condition of water and sediment quality.

\section{Methodologies}

\textbf{Data Collection:}
The collected dataset consists of 46 features and 271 entries.
There is a high percentage (around 70\%) of missing data within the dataset. There are several reasons associated with this issue. Firstly, some non-heavy elements such as transition metal, silver, mercury, and beryllium are usually below detection. For example, the elements in Australian ports do not typically attempt to detect such a metal. Secondly, the data collected from different resources typically do not contain the same features because there are no established standards. The set of detected materials across studies might vary due to the scope differences.
Thirdly, some studies might even contain organic elements detection, while other datasets might not. Therefore, we conducted a feature selection process that focuses on necessary and meaningful features. There are 25 features with 271 entries remaining in the refined dataset, and the missing rate drops to around 53\%.

\subsubsection{Data Labelling}

\begin{figure*}
\centerline{\includegraphics[width=\linewidth]{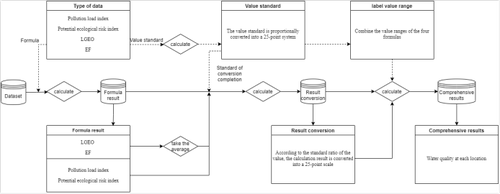}}
\caption{The overall data labelling process.\label{fig2}
}
\end{figure*}

We synthesized the target variable by utilsing four pollution indicators: Igeo, EF, pollution load index (PLI) and potential ecological risk index (PER). The indicators are used to assess water quality based on various types of water sediment.
Fig. \ref{fig2} illustrates the overall process of data labelling. 
Firstly, we compute the four indicators in accordance with their specifications. 
Among these results, Igeo and EF are calculated for each element in the water in the area, while PLI and PER are a comprehensive evaluation of all elements in the area.
Because the number of levels among each indicator is not the same, it is necessary to systematically merge the label intervals among indicators.

For geoaccumulation indices (Igeo), the merger criteria is as follows:
\begin{equation}
   f_\text{score}(x_\text{Igeo}) =
    \begin{cases}
        0 &\text{ if $x_\text{Igeo} < 0$},\\
        5x &\text{ if $0 \le x_\text{Igeo} \le 5$},\\
        25 &\text{ if $x_\text{Igeo} > 5$}.
    \end{cases}
\end{equation}
For the enrichment factors (EF), the merger criteria is as follows:
\begin{equation}
   f_\text{score}(x_\text{ef}) =
    \begin{cases}
        0 &\text{ if $x_\text{ef} < 2$},\\
        \frac{25(x-2)}{38} &\text{ if $2 \le x_\text{ef} \le 40$},\\
        25 &\text{ if $x_\text{ef} > 40$}.
    \end{cases}
\end{equation}
For the pollution load index (PLI), the merger criteria is as follows:
\begin{equation}
   f_\text{score}(x_\text{pli}) =
    \begin{cases}
        0 &\text{ if $x_\text{pli} < 1$},\\
        \frac{25(x-1)}{4} &\text{ if $1 \le x_\text{pli} \le 5$},\\
        25 &\text{ if $x_\text{pli} > 5$}.
    \end{cases}
\end{equation}
For the pollution load index (PER), the merger criteria is as follows:
\begin{equation}
   f_\text{score}(x_\text{per}) =
    \begin{cases}
        0 &\text{ if $x_\text{per} < 40$},\\
        \frac{25(x-40)}{280} &\text{ if $40 \le x_\text{per} \le 320$},\\
        25 &\text{ if $x_\text{per} > 320$}.
    \end{cases}
\end{equation}

\begin{table}
\caption{The four levels of criteria}
\label{table1}
\centering
\resizebox{\linewidth}{!}{
\begin{tabular}{c|c|c|c}
\hline
\multicolumn{2}{c|}{\textbf{Original Standard}} & \multicolumn{2}{c}{\textbf{Artificial Merger Criteria}}\\
\hline
Igeo\textless 0 & Unpolluted & Igeo\textless 0 & Unpolluted\\
\hline
0\textless Igeo\textless 1 & \makecell[c]{Unpolluted to\\ moderately polluted} & 0\textless Igeo\textless 1 & Light pollution\\
\hline
1\textless Igeo\textless 2 & Moderately polluted & \multirow{2}{*}{1\textless Igeo\textless 3} & \multirow{2}{*}{Moderate pollution}\\
\cline{1-2}
2\textless Igeo\textless 3 & \makecell[c]{Moderately to\\ strongly polluted} & & \\
\hline
3\textless Igeo\textless 4 & Strongly polluted & \multirow{3}{*}{3\textless Igeo} & \multirow{3}{*}{Heavy pollution}\\
\cline{1-2}
4\textless Igeo\textless 5 & Strongly polluted & & \\
\cline{1-2}
5\textless Igeo & Extremely polluted & & \\
\hline
\end{tabular}
}
\end{table}

\subsubsection{Data Imputation}

The data we collected has a large percentage of missing data, around 53\%, which is problematic for the data-driven machine learning model. Therefore, a well-performing data imputation method is necessary to standardise and clean the to enhance the performance of machine learning models. We design an experiment to examine the efficiency of different imputation models, which includes: (i) simple imputation, (ii) k-nearest neighbour (KNN) imputation, (iii) singular value decomposition (SVD) imputation, and (iv) iterative imputation.

In our experiment, we select a dataset as the target domain. Then, experimentally, we evaluated the performance of various imputation methods.  We performed the imputation methods under different missing rates and repeated the experiment ten times to obtain a statistically significant result. We randomly drop values with a missing rate ranging from 0.35 to 0.65, with an increment step of 0.05. Then we compared the mean of the symmetric mean absolute value (SMAPE) between imputed and the actual values. The highest performing imputation method is selected in this research.

Table~\ref{table:mean-svd} and~\ref{table:knn-iterative} illustrate the performance differences in-between several data imputations methods.
Simple imputation refers to using the mean across each feature column to fill missing values.
Data standardisation refers to rescaling the value of each feature to the same scale to let all features contribute equally to the developed model. Several ways to standardise include min and max normalisation, z-score standardisation, and centralisation. Standardisation rescales the value into zero means and one unit standard deviation. Normalisation is to rescale the value into 0 and 1. Centralisation is to rescale the value to be centred at 0.

In classification and clustering algorithms, z-score standardisation performs when the distance is calculated to measure the similarity, or dimensionality reduction technologies such as principal component analysis (PCA) are applied. In our dataset, the models we propose to develop, such as support vector machines (SVM) and KNN, highly rely on distance calculation, so we adopt the z-score standardisation technique.

\subsubsection{Data Augmentation}

Our labelled dataset is highly imbalanced, with label ‘A’ accounting for 43$\%$, label ‘B’ accounting for 51$\%$, and label ‘C’ accounting for 6$\%$. In an imbalanced dataset, the predicting accuracy (the number of correctly predicted samples / the total number of samples) will become ineffective. This is because the positive samples in an imbalanced dataset occupy a large percentage, and the accuracy score will become high even if none of the negative samples is successfully predicted. When the ratio of two groups of samples exceeds 4:1, the imbalance problem will be severe.

There are three standard methods for dealing with an imbalanced dataset: (i) resampling, (ii) over-sampling, and (iii) under-sample. Our original data entries are small, so we choose over-sampling, synthesising data entries for the minority label classes. This process is also known as data augmentation. The most straightforward augmentation technique is picking a small number of samples at random, then making a copy and adding it to the whole sample. However, if the feature dimension of the data is small, simple extraction and replication will lead to over-fitting easily. We adopted a new augmentation method called Synthetic Minority Over-sampling Technique (SMOT). SMOT is to find K numbers of neighbours in P dimensions and then multiply each index of the K neighbours by a random number between 0 and 1 to form a new minority class sample. SMOT can introduce some noise to the synthesised samples to avoid the problem of overfitting.

\subsubsection{Machine Learning Model Developing}

The machine learning models we adopted in this research are Logistic Regression, Naive Bayes, Decision Tree, KNN, SVC and MLPClassifier. Logistic regression is a linear regression plus a sigmoid function that can convert numeric prediction into categorical outcomes. Naive Bayes classifier is a classification technique that uses Bayes' Theorem as its underlying theory. It assumes that all the predictors are independent; that is to say, all features are unrelated and do not have any correlation. KNN is the k-nearest neighbour classifier. It calculates the distances between the data entry to be predicted and other data entries, then votes for the most frequent label among k closest number of data entries to predict the label of the target entry ($k$ is a hyperparameter that needs to be tuned during model training). SVC stands for support vector classifier. There is no required assumption on the shape of features, and no parameters need to be tuned. It generates the 'best fit' hyperplane that divides or categorizes the samples. MLPClassifier is a multi-layer perceptron (MLP) that trains a neural network using a backpropagation algorithm.

Besides MLPClassifier, we build a fully connected deep neural network (DNN) and tune the hyperparameters to find a DNN model with the highest predicting accuracy. DNN is an artificial neural network with many hidden layers between its input and output layers. The number of neurons in the input and output layer is the same as the number of data entries and different labels in the dataset. A weighted sum plus a bias is applied to all the neurons in the previous layer. A non-linear function is used to change its linearity. Then the calculated value works as the input value of neurons in the next layer. The weight of each neuron will not be updated through the backpropagation algorithm until the loss function is minimized. Then the latest weight, together with other parameters, form the architecture of the DNN model.

Parameter tuning is the most crucial task during DNN model development. In this project, we use predicting accuracy to examine the efficiency of different parameters. Typical parameters include the number of hidden layers, the number of neurons in each layer, the activation function, the dropout rate, the batch normalization, the epoch, and the gradient descent method. We first tune the number of hidden layers and then use the best number of hidden layers to tune the number of neurons in each layer. Then, we tune the dropout rate and evaluate batch normalization's effectiveness.

\subsubsection{Model Comparison}

Logistic regression is one of the straightforward and easy-to-interpret algorithms in machine learning, so it is used as the benchmark model in our comparison. Prediction accuracy and F1-score will be compared between those models. Details of machine learning model accuracy evaluation methods will be discussed in the subsequent section.

\subsection{Data Collection}

The dataset used in this research was collected from some authoritative websites and combined by the following six datasets from various sources.

\begin{enumerate}
    \item Dataset I was collected by Jahan and Strezov (2018). Using Ekman grab sampler, they collected sediment samples from three different locations at each of the six ports of Sydney, Jackson port, Botany port, Kembla port, Newcastle port, Yamba port and Eden port, and obtained the content data of 42 different substances in sediments at different sampling points\cite{3}.
    \item Our Dataset II comes from Perumal et al. (2019), which includes surface sediments from 24 different locations in the areas affected by different anthropogenic activities in the coastal area of Thondi using Van Veen grab surface sampler, and measured the grain size, organic substance and heavy metal concentration of surface sediment samples \cite{12}.
    
    \item Dataset III was collected by Fan et al. in 2019. Using the Bottom Sediment Grab Sampler, they collected 70 surface sediment samples in Luoyuan Bay, northeast coast of Fujian Province, China, and measured the concentrations of eight heavy metals V, Cr, Co, Ni, Cu, Zn, Cd, and Pb in the sediment samples\cite{13}.
    
    \item Dataset IV was collected by Constantino et al. (2019) at nine different sampling points in the Central Amazon basin. During the dry season, they collected sediment samples with a 30 by 30 cm Ekman-Birge dredger and measured the concentration of different metals in sediment samples\cite{14}.
    
    \item Dataset V was collected in a polygon along the Brazilian continental slope of the Santos Basin, known as the Sao Paulo Bight, in an arc area on the southern edge of Brazil. They used the box corer device to collect sediment samples at thirteen sampling points, and measured the concentration of some metals in surface sediments at these 13 sampling points and the concentration of metals in sediments at different depths at some sampling points, to obtain Dataset \uppercase\expandafter{\romannumeral5}\cite{15}.

    \item In 2016, Qiu et al. used the rotary drilling method to collect sediment cores from the coastal area of Jiangsu Province, China, and measured the concentration of metals in 80 samples at different depths in the cores to obtain our Dataset VI.
\end{enumerate}

\subsection{Evaluation}\label{sec:evaluation}

\textbf{Data Imputation:} Evaluating the appropriate data imputation approach  is an essential process to address the issue with missing data which is common among sediment data. We have selected four state-of-the-art approaches as the candidate methods. The designed experiment calculated the average accuracy of different imputation methods under different missing rates. SMAPE is used to evaluate the models' performance. Using SMAPE instead of MAE or RMSE helps to account for the magnitude differences between features. 
In addition, accuracy and F1-score are used to examine the performance of the models. Various performance metrics can be used to evaluate the performance of models. The most basic one is the overall accuracy. The calculation divides the number of correct predictions by the total number of predictions. Accuracy is a direct metric indicating if models can predict the result correctly.  However, accuracy is unreliable in the problem of the classification of highly skewed data. Failure to classify the data to minority classes will not affect obtaining high accuracy. F1-score is introduced in the evaluation process to improve the evaluation of models, which considers both precision and recall. Precision refers to the percentage of samples classified as a particular class among all samples predicted to be classified as a particular class. In contrast, recall refers to the percentage of samples predicted to be classified as a particular class among all samples classified as a specific class. F1-score is given by
\begin{equation}
    2 (precision \times recall) / (precision + recall),
\end{equation}
which comprehensively considers precision and recall and can be used to comprehensively evaluate the prediction accuracy of a machine learning model.  Therefore, precision, recall, and F1-score are used to comprehensively judge the performance of different models.

The accuracy of the DNN model in different training episodes might exhibit different results, which is because of the randomness during neural network training. Randomness happens due to several reasons. The first reason is that the weight from the upper to the lower neurons will be initialized randomly. Second, there is also the randomness from utilizing data imputations methods to fill missing data features. Our dataset is relatively small; therefore, the accuracy can vary substantially across runs. Therefore, we provide statistically significant results by running our experiments multiple times to obtain the average of multiple training results.

\section{Results}

\subsection{Data Imputation Results}

\begin{table*}[tb]
\caption{The SMAPE score (mean and standard deviation) by using Mean Imputation and SVD Imputation.\label{table:mean-svd}}
\begin{tabular}{@{}cccccccc@{}}
\toprule
    \multicolumn{1}{c}{\multirow{2}{*}{\textbf{Imputation Method}}}                                                  & \multicolumn{7}{c}{\textbf{Data Missing Rate}} \\ \cmidrule(l){2-8} 
                           & \textbf{0.35} & \textbf{0.4} & \textbf{0.45} & \textbf{0.5} & \textbf{0.55} & \textbf{0.6} & \textbf{0.65} \\ \midrule
Mean Imputation            & 115.42±3.45   & 111.44±3.11  & 112.06±4.81   & 111.58±6.87  & 108.61±7.24   & 110.51±5.47  & 109.02±6.24   \\
SVD Imputation             & 112.63±4.80   & 114.08±4.08  & 112.91±5.99   & 111.97±4.06  & 114.57±7.96   & 106.98±6.61  & 106.30±6.68   \\ \bottomrule
\end{tabular}
\end{table*}

\begin{figure}[tb]
\centerline{\includegraphics[width=18.5pc]{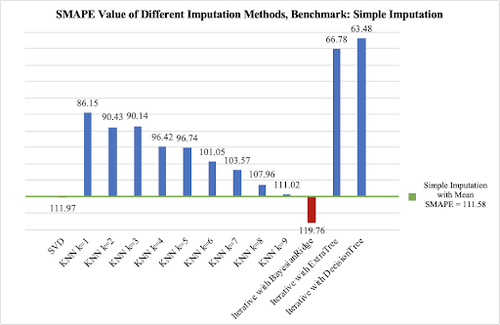}}
\caption{SMAPE value of different imputation methods.\label{fig3}}
\end{figure}

The SMAPE value of different missing data imputation methods is shown in Table \ref{table:mean-svd}, illustrated by its mean and standard deviation. Table \ref{table:knn-iterative} shows the iterative imputation using different tree algorithms with a different number of trees. Fig. \ref{fig3} gives a clear comparison of different imputation methods with simple imputation as the benchmark. The green line indicates that the SMAPE of simple imputation is 111.58. Any column above the green line means that the method performs better than simple imputation and vice versa.

\begin{table*}[tb]
\caption{The SMAPE score (mean and standard deviation) by using kNN Imputation and Iterative Imputation \label{table:knn-iterative}}
\resizebox{\linewidth}{!}{%
\begin{tabular}{@{}ccccccccc@{}}
\toprule
\multicolumn{2}{c}{\multirow{2}{*}{\textbf{Imputation Method}}}                                                  & \multicolumn{7}{c}{\textbf{Data Missing Rate}} \\ \cmidrule(l){3-9} 
              \multicolumn{2}{c}{}                                 & \textbf{0.35}                & \textbf{0.4}                 & \textbf{0.45}                & \textbf{0.5}                 & \textbf{0.55}                & \textbf{0.6}                 & \textbf{0.65}                \\ \midrule
\multirow{9}{*}{kNN Imp.}      & \multicolumn{1}{c|}{$k=1$}            & 75.09±7.53  & 80.54±5.97  & 85.79±4.37  & 86.15±5.49  & 82.36±6.75  & 82.92±9.38  & 85.51±5.33  \\
                                      & \multicolumn{1}{c|}{$k=2$}            & 80.73±5.79  & 84.13±6.35  & 86.87±4.49  & 90.43±5.49  & 87.96±6.76  & 87.88±5.95  & 94.30±6.75  \\
                                      & \multicolumn{1}{c|}{$k=3$}            & 85.12±9.81  & 88.97±5.47  & 86.06±7.86  & 90.14±5.22  & 91.54±6.96  & 96.43±3.02  & 98.98±5.04  \\
                                      & \multicolumn{1}{c|}{$k=4$}            & 84.14±4.57  & 89.13±5.05  & 93.15±5.92  & 96.42±5.61  & 96.95±7.79  & 99.70±5.14  & 103.67±5.03 \\
                                      & \multicolumn{1}{c|}{$k=5$}            & 90.77±7.31  & 98.49±10.10 & 97.63±4.98  & 96.74±6.74  & 98.85±4.60  & 102.70±5.32 & 104.09±4.83 \\
                                      & \multicolumn{1}{c|}{$k=6$}            & 96.79±4.43  & 97.39±3.65  & 101.02±7.03 & 101.05±7.35 & 104.48±5.39 & 106.65±6.29 & 108.54±4.63 \\
                                      & \multicolumn{1}{c|}{$k=7$}            & 93.49±5.92  & 96.89±5.93  & 101.57±4.79 & 103.57±9.84 & 112.02±6.40 & 109.42±7.54 & 107.64±7.20 \\
                                      & \multicolumn{1}{c|}{$k=8$}            & 97.73±5.80  & 101.13±3.51 & 104.13±3.31 & 107.96±5.36 & 111.33±9.96 & 106.18±5.08 & 108.25±7.20 \\
                                      & \multicolumn{1}{c|}{$k=9$}            & 100.12±8.11 & 100.09±8.82 & 109.13±9.04 & 111.02±4.35 & 109.28±3.34 & 108.88±8.14 & 105.98±8.01 \\ \midrule
\multirow{3}{*}{Iterative Imp.} & \multicolumn{1}{c|}{Bayesian Ridge} & 114.92±7.78 & 116.46±6.56 & 122.05±5.15 & 119.76±6.09 & 123.83±5.28 & 123.91±6.43 & 123.25±7.18 \\
                                      & \multicolumn{1}{c|}{Decision Tree}  & 60.11±5.19  & 62.22±3.31  & 64.27±3.68  & 66.78±6.40  & 68.75±5.10  & 70.34±9.64  & 80.04±9.06  \\ \midrule
\multirow{5}{*}{Extra Tree} & \multicolumn{1}{c|}{$n=1$} & 58.10±5.79 & 58.65±3.53 & 62.17±4.74 & 64.13±3.44 & 66.45±6.01 & 68.91±7.23 & 72.81±5.62 \\
 & \multicolumn{1}{c|}{$n=5$} & 60.25±4.00 & 61.84±2.37 & 58.40±33,35 & 65.05±5.44 & 63.36±3.36 & 66.88±3.27 & 70.91±7.69 \\
 & \multicolumn{1}{c|}{$n=10$} & 57.40±3.65 & 59.05±3.91 & 62.78±4.20 & 63.76±4.47 & 62.69±4.63 & 68.63±4.26 & 70.41±8.61 \\
 & \multicolumn{1}{c|}{$n=20$} & 62.46±2.72 & 64.33±5.98 & 64.98±5.55 & 66.38±4.25 & 67.52±4.89 & 70.20±3.13 & 77.58±5.22 \\
 & \multicolumn{1}{c|}{$n=50$} & 62.13±5.94 & 65.99±6.18 & 65.94±4.49 & 68.34±4.50 & 71.76±3.32 & 71.00±5.41 & 76.57±6.59  \\ \bottomrule
\end{tabular}
}
\end{table*}

\begin{table}[tb]
\caption{The mean and standard deviation of SMAPE in ExtraTree with different tree numbers}
\label{table7}
\centering
\resizebox{\linewidth}{!}{%
\begin{tabular}{c|c|c|c|c|c|c|c}
\hline\hline
\textbf{\makecell[c]{Missing \\Rate (\%)}} & \textbf{0.35} & \textbf{0.4} & \textbf{0.45} & \textbf{0.5} & \textbf{0.55} & \textbf{0.6} & \textbf{0.65}\\
\hline
\multicolumn{8}{c}{\textbf{Iterative Imputation with Extra Tree Estimator}}\\
\hline
n=1 & \makecell[c]{58.10\\±5.79} & \makecell[c]{58.65\\±3.53} & \makecell[c]{62.17\\±4.74} & \makecell[c]{64.13\\±3.44} & \makecell[c]{66.45\\±6.01} & \makecell[c]{68.91\\±7.23} & \makecell[c]{72.81\\±5.62}\\
\hline
n=5 & \makecell[c]{60.25\\±4.00} & \makecell[c]{61.84\\±2.37} & \makecell[c]{58.40\\±33,35} & \makecell[c]{65.05\\±5.44} & \makecell[c]{63.36\\±3.36} & \makecell[c]{66.88\\±3.27} & \makecell[c]{70.91\\±7.69}\\
\hline
n=10 & \makecell[c]{57.40\\±3.65} & \makecell[c]{59.05\\±3.91} & \makecell[c]{62.78\\±4.20} & \makecell[c]{63.76\\±4.47} & \makecell[c]{62.69\\±4.63} & \makecell[c]{68.63\\±4.26} & \makecell[c]{70.41\\±8.61}\\
\hline
n=20 & \makecell[c]{62.46\\±2.72} & \makecell[c]{64.33\\±5.98} & \makecell[c]{64.98\\±5.55} & \makecell[c]{66.38\\±4.25} & \makecell[c]{67.52\\±4.89} & \makecell[c]{70.20\\±3.13} & \makecell[c]{77.58\\±5.22}\\
\hline
n=50 & \makecell[c]{62.13\\±5.94} & \makecell[c]{65.99\\±6.18} & \makecell[c]{65.94\\±4.49} & \makecell[c]{68.34\\±4.50} & \makecell[c]{71.76\\±3.32} & \makecell[c]{71.00\\±5.41} & \makecell[c]{76.57\\±6.59}\\
\hline
\hline
\end{tabular}
}
\end{table}

\subsection{Deep Learning Model Tuning Results}

\subsubsection{Tuning the Number of Hidden Layers}

\begin{figure*}[tb]
\centering
\includegraphics[width=\linewidth]{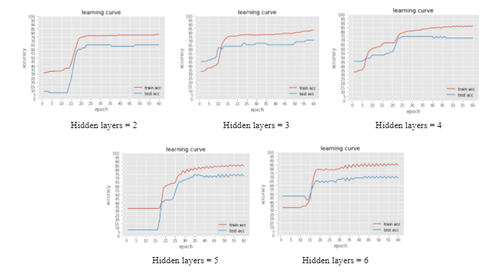}
\caption{The accuracy of the train and test dataset under different hidden layers numbers (epoch=60).\label{fig4}}
\end{figure*}

Fig. \ref{fig4} illustrates the prediction accuracy of the train and test dataset under different numbers of hidden layers ranging from 2 to 6, increased by 1. The test accuracy with two hidden layers equals 0.65, increases to 0.75 when hidden layers are 4 and 5, then drops to 0.70 with six hidden layers. The training accuracy for 4 and 5 hidden layers is 0.87 and 0.85 separately. Hence, the five hidden layers prevail over four hidden layers due to less overfitting.

\subsubsection{Tuning the Number of Neurons in Each Layer}

\begin{figure*}[tb]
\includegraphics[width=\linewidth]{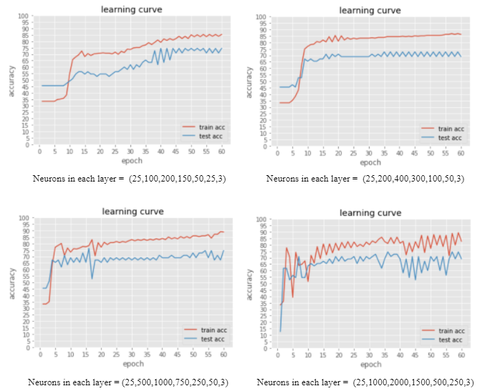}
\caption{The accuracy of train and test dataset under different neurons numbers (hidden layers=5).\label{fig5}}
\end{figure*}

We choose the neural network structure with 5 hidden layers to tune the number of neurons in each hidden layer. From Fig. \ref{fig5} we can see that model with small neuron numbers (25, 100, 200, 150, 50, 25, 3) has a low prediction accuracy, and with large neuron numbers (25, 1000, 2000, 1500, 500, 250, 3) has an unstable performance during each epoch. Models with neuron numbers (25, 200, 400, 300, 100, 50, 3) and (25, 500, 1000, 750, 250, 50, 3) have similar performance but scholars tend to select less complex models in general. So we choose neurons in each layer equal (25, 200, 400, 300, 100, 50, 3).

\subsubsection{Tuning the Dropout Rate}

\begin{figure}[tb]
\centerline{\includegraphics[width=18.5pc]{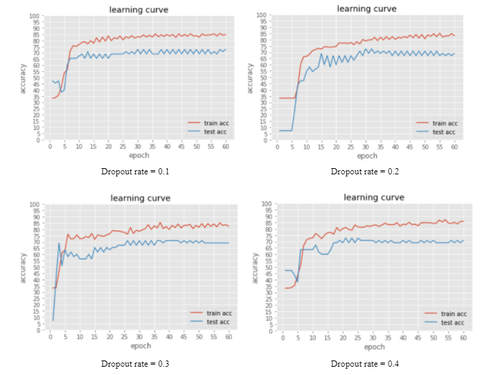}}
\caption{The accuracy of train and test dataset under different dropout rate (hidden layers=5, Neurons in each layer = (25,200,400,300,100,50,3)).\label{fig6}}
\end{figure}

From the previous tuning process, we find an overfitting problem during model development. We tune the dropout rate ranging from 0.1 to 0.4, which is increased by 0.1 to lessen the accuracy gap between train and test datasets. Fig. \ref{fig6} shows that the performances difference between different dropout rates is similar, but the 0.2 dropout rate has a slight advantage around 30 epochs, where the training accuracy is 80, and the testing accuracy is 0.75.

\subsubsection{Tuning the Batch Normalization}

\begin{figure}[tb]
\centerline{\includegraphics[width=18.5pc]{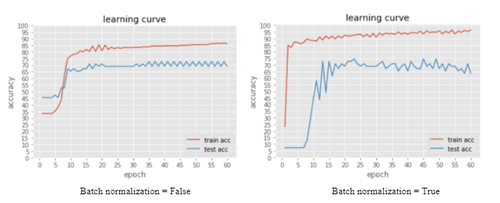}}
\caption{The accuracy of train and test dataset under different batch normalization (hidden layers=5, Neurons in each layers = (25,200,400,300,100,50,3)).\label{fig7}}
\end{figure}

Batch normalization is another technique used to solve the overfitting problem. The criterion is whether to adopt batch normalization after each hidden layer or not. The left-hand side graph in Fig. \ref{fig7} indicates not using batch normalization, and the right-hand side graph indicates using batch normalization. Using batch normalization performs worse than not to use.

\subsubsection{Tuning Both Dropout Rate and Batch Normalization}

\begin{figure}
\centerline{\includegraphics[width=18.5pc]{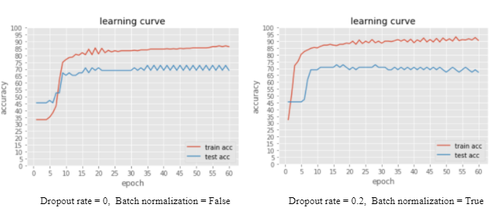}}
\caption{The accuracy of train and test dataset under different dropout rate and batch normalization (hidden layers=5, Neurons in each layers = (25,200,400,300,100,50,3)).\label{fig8}}
\end{figure}

We had attempted to tune dropout and batch normalization jointly. The left-hand side graph in Fig. \ref{fig8} indicates not using dropout and batch normalization, and the right-hand side graph indicates using 0.2 dropout rate (the best rate from the previous tuning) and batch normalization. Similarly, using both techniques performs worse than not using.

\begin{figure}
\centerline{\includegraphics[width=18.5pc]{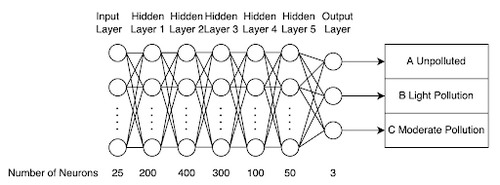}}
\caption{Architecture of DNN model.\label{fig9}}
\end{figure}

\textbf{Architecture of the Final DNN Model:}
From above model tuning, we finally choose the DNN model with one input layer, five hidden layers, and one output layer. The number of neurons in each layer is (25, 200, 400, 300, 100, 50, 3). There is no dropout or batch normalization in this model. The architecture is shown in Fig. \ref{fig9}.

\subsection{Model Development Results}

\begin{figure}[tb]
\centerline{\includegraphics[width=18.5pc]{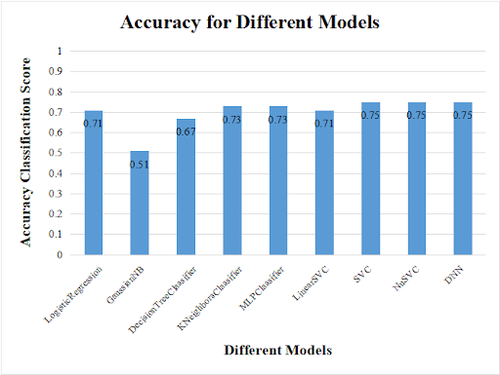}}
\caption{Accuracy for different models.\label{fig10}}
\end{figure}

\begin{figure}[tb]
\centerline{\includegraphics[width=18.5pc]{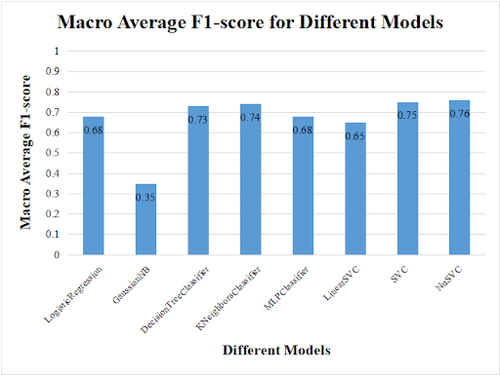}}
\caption{Macro average F1-score for different models.\label{fig11}}
\end{figure}

The values of the accuracy score corresponding to different trained models are shown in  Fig. \ref{fig10}. It can be seen that SVC, NuSVC and DNN models have the highest performance on this metric, and their corresponding accuracy classification score is 0.75.
The performance of each machine learning model on F1-score is shown in the Fig. \ref{fig11}. It can be seen that the NuSVC model has the highest performance on this metric, and its corresponding F1-score is 0.76.
As mentioned earlier, the above metrics were used to evaluate the prediction accuracy of a model. The larger the value of the above metrics, the higher the accuracy of a model. The following Table \uppercase\expandafter{\romannumeral8} shows the performance of different models on the metrics of ‘accuracy’, ‘macro average precision’, ‘macro average recall’, and ‘macro average F1-score’. The performance of the NuSVC model is the highest.

\begin{table}[tb]
\renewcommand{\arraystretch}{1.3}
\caption{Performance comparison of models}
\label{table8}
\centering
\setlength{\tabcolsep}{0.8pt}
\begin{tabular}{c|c|c|c|c}
\hline
\textbf{Models / Metrics} & \textbf{Accuracy} & \textbf{\makecell[c]{Macro \\Average\\ Precision}} & \textbf{\makecell[c]{Macro \\Average\\ Recall}} & \textbf{\makecell[c]{Macro \\Average\\ F1-score}}\\
\hline
LogisticRegression & 0.710 & 0.660 & 0.720 & 0.680\\
\hline
GaussianNB & 0.510 & 0.360 & 0.370 & 0.350\\
\hline
DecisionTreeClassifier & 0.670 & 0.770 & 0.700 & 0.730\\
\hline
KNeighborsClassifier & 0.730 & 0.720 & 0.810 & 0.740\\
\hline
MLPClassifier & 0.730 & 0.710 & 0.660 & 0.680\\
\hline
LinearSVC & 0.710 & 0.650 & 0.650 & 0.650\\
\hline
SVC & 0.750 & 0.750 & 0.750 & 0.750\\
\hline
NuSVC & 0.750 & 0.730 & 0.820 & 0.760\\
\hline
DNN & \makecell[c]{0.748\\±0.018} & N/A & N/A & N/A\\
\hline
\end{tabular}
\end{table}

\section{Discussion}

\subsection{Labelling Process}

\begin{figure}
\centerline{\includegraphics[width=18.5pc]{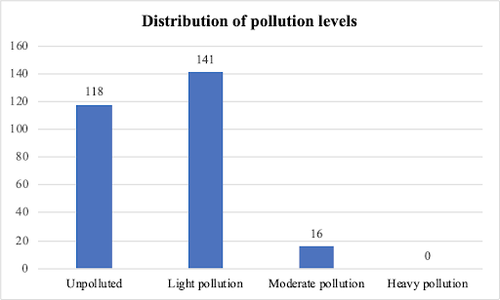}}
\caption{The distribution of labelling pollution levels.\label{fig12}}
\end{figure}

Labelling the data with the pollution level is essential in this research. The labelling standard was made for research, so it is essential to consider its reliability. This is because our research result will be meaningless if the label is unreliable. No specific performance metrics can be chosen to evaluate that process about labelling the level of pollution, but the distribution of the level of pollution is consistent with our background research. According to Fig. \ref{fig12}, the distribution of different pollution levels mainly focuses on polluted and light pollution, which accords to real-life situations. There is no severe pollution which is reasonable since the government will control that situation. In addition, the algorithm we chose to calculate the level of pollution reflects the pollution degree values the metal concentration of sediments more.

From our research, there are a few detailed applications to various areas. Firstly, a practical system to evaluate the level of pollution is achieved. As mentioned before, it is believed that the system to estimate the level of pollution using indexes is entirely usable for a simple estimation. Although the guideline of labelling different data into four levels of pollution was created in this research, its reasonability is solid. The comprehensive evaluation of water quality is complicated, and it will consider many factors, including some compounds that were not included in this research\cite{17}. The guideline established in this research followed the process of water quality guidelines. In many situations, the measurement of all various compounds cannot be done due to the complexity. Our guideline, which can generate pollution labels for sediment samples by the mental concentration in water sediments, is a good choice for simple and preliminary evaluation.

\subsection{Missing Data Imputation}

Besides, we experimented to find the best methods for missing data imputation. The experiment result of which method will be better for imputation can be referenced by experiments with a small dataset having a large percentage of missing data. Similar situations can reference the result. In the case of this research, the rate of missing data is about 53\% which is too high to take standard methods of missing data imputation such as filling the missing data with mean, especially when the dataset is relatively small.

As the missing rate in our dataset is around 53\%, we focus on the SMAPE values of different missing data imputation methods under the 0.50 missing data rate. A lower score in SMAPE value implies a better imputation performance. All the imputation methods perform better than simple imputation except iterative imputation with BayesianRidge and SVD. SVD performs slightly worse than simple imputation. Iterative imputation with extra trees has the best performance. Notably, the number of trees (n) in extra tree regression can be tuned as well, so we tried n equals 1, 5, 10, 20, 50, and the outcome in Section \uppercase\expandafter{\romannumeral6}-A clearly shows that the SMAPE of iterative imputation with extra trees is the lowest when n equals ten. In conclusion, the selected imputation method in this project is iterative imputation with the extra tree algorithm (n=10) as the estimator.

\subsection{Model Developing}

The most significant achievement of this research is the classification model used to predict pollution, and it has many applications to real-world problems. Using this model, people can predict pollution levels using the concentrations of different metals in marine sediments. A practical application of this model is that this model can be used in protecting aquatic ecosystems since the protection and policies start from reorganization. The model can also be used to monitor future water contamination. Future water contamination monitoring can address the detection of priority mixtures\cite{18}.

From our result, it can be deduced from the algorithm calculating the pollution level and variable importance plot that Fe, Cu, Zn, Co, Mn, Se and Ni are the metals that most affect the quality of sediments. The practical significance of this result is that the government will know the concentration of what kind of metals in sediment should be strictly controlled.

\subsection{Selected Best Model}

According to Section \uppercase\expandafter{\romannumeral6}-C, the accuracy scores of NuSVC and DNN models are about 0.75. In other words, the percentages of correct prediction classification of NuSVC and DNN models are similar. However, the F1-score of DNN model is unavailable, as is mentioned in Section \uppercase\expandafter{\romannumeral4}-D, it is difficult to compare the performance of NuSVC and DNN through that. Since the available dataset collected in this research is too tiny, the difference in training speed between NuSVC and DNN models is not apparent. In addition, there is no complex hyperparameter adjustment process in training the NuSVC model compared with DNN model.

In conclusion, the NuSVC model is the best model for this research. It can separate points belonging to different categories to the greatest extent to ensure classification accuracy. Although the support vector machine model was initially proposed for two kinds of classification problems, the NuSVC model can also be used for multi-class classification problems that need to be solved in this research by adequately adjusting the model parameters. In addition, the NuSVC model is suitable for high-dimensional datasets, which is suitable for our high dimensional dataset. Unfortunately, this model is difficult to visualize, making it impossible for us to provide a visual model to users, as mentioned above.

\section{Limitations and Future Works}

Water sediment is an essential part of the ecological environment, and its physical and chemical properties will affect biological integrity. Therefore, investigating and studying water sediments is one of the ways to detect the quality of the water environment. This research uses machine learning technology to evaluate the quality of the water-sediment samples taken to evaluate the quality of seawater samples at different locations and depths and provide adequate information to evaluate the quality of the nearby water environment.

From a technical perspective, machine learning is used in research to build models based on collected data samples to provide compelling predictions or decision-making when applied to new samples. Analysing seawater samples using machine learning can evaluate and predict water quality by analysing water sediments' state, composition, and content.

These prediction results, even machine learning models, will be provided to environmental experts and related departments for further analysis in their professional fields to have further analysis of water sediments, such as tracing its source or suggesting treatments and improvement methods, and ultimately play a positive role in the prevention and treatment of water pollution.

\subsection{Limitations}

At present, the dataset we collected may affect the accuracy of model training. The reasons can be summarised as the following:

\begin{enumerate}
\item[{\it a)}]{\it Few Effective Data Sources}

When collecting datasets related to the quality of water sediments, we found that some high-quality datasets were private or protected. At the same time, some sediment samples of some datasets only contain some data about biological pollution factors, and the concentration of metal elements is not detected. Therefore, these datasets cannot be applied to our research.

\item[{\it b)}]{\it Missing Data}

As the datasets are collected by different researchers, some datasets might contain incomplete data, resulting in missing items when these datasets are merged. Although some missing data processing measures have been taken, they may still affect the accuracy of the final results.

\item[{\it c)}]{\it Differential Concentration of Values}

The accuracy and magnitude of measurements are different across studies (due to the differences in their research focus or equipment used). The inconsistency might introduce artificial inaccuracy across datasets, which could affect the accuracy of the final result.
\end{enumerate}

\section{Conclusion}
We propose a unified framework for introducing the predictive capability of modern machine learning techniques into water and sediment analysis.
Our framework provide a systemic approach to evaluate the most appropriate data imputation methods to tackle the data scarcity and missing data issue, which are typical in existing studies.
Our final model archives state-of-the-art performance across other models to classify water pollution level.

\bibliographystyle{ieeetr}
\bibliography{bibfile}

\begin{thebibliography}{10}

\bibitem{1}
J.~Fu, X.~Hu, X.~Tao, H.~Yu, and X.~Zhang, ``Risk and toxicity assessments of
  heavy metals in sediments and fishes from the yangtze river and taihu lake,
  china,'' {\em Chemosphere}, vol.~93, no.~9, pp.~1887--1895, 2013.

\bibitem{2}
J.~Chen, M.~Liu, N.~Bi, Y.~Yang, X.~Wu, D.~Fan, and H.~Wang, ``Variability of
  heavy metal transport during the water–sediment regulation period of the
  yellow river in 2018,'' {\em Science of The Total Environment}, vol.~798,
  p.~149061, 2021.

\bibitem{3}
S.~Jahan and V.~Strezov, ``Comparison of pollution indices for the assessment
  of heavy metals in the sediments of seaports of nsw, australia,'' {\em Marine
  Pollution Bulletin}, vol.~128, pp.~295--306, 2018.

\bibitem{4}
X.-D. Li, L.~Xin, W.-T. Rong, X.-Y. Liu, W.-A. Deng, Y.-C. Qin, and X.-L. Li,
  ``Effect of heavy metals pollution on the composition and diversity of the
  intestinal microbial community of a pygmy grasshopper (eucriotettix
  oculatus),'' {\em Ecotoxicology and Environmental Safety}, vol.~223,
  p.~112582, 2021.

\bibitem{5}
C.~Zhang, Q.~Liu, B.~Huang, and Y.~Su, ``Magnetic enhancement upon heating of
  environmentally polluted samples containing haematite and iron,'' {\em
  Geophysical Journal International}, vol.~181, no.~3, pp.~1381--1394, 2010.
\newblock cited By 27.

\bibitem{6}
C.~Zhang, Q.~Qiao, J.~D. Piper, and B.~Huang, ``Assessment of heavy metal
  pollution from a fe-smelting plant in urban river sediments using
  environmental magnetic and geochemical methods,'' {\em Environmental
  Pollution}, vol.~159, no.~10, pp.~3057--3070, 2011.
\newblock Nitrogen Deposition, Critical Loads and Biodiversity.

\bibitem{7}
N.~U. Saher and A.~S. Siddiqui, ``Comparison of heavy metal contamination
  during the last decade along the coastal sediment of pakistan: Multiple
  pollution indices approach,'' {\em Marine Pollution Bulletin}, vol.~105,
  no.~1, pp.~403--410, 2016.

\bibitem{8}
J.~Yang, L.~Chen, L.-Z. Liu, W.-L. Shi, and X.-Z. Meng, ``Comprehensive risk
  assessment of heavy metals in lake sediment from public parks in shanghai,''
  {\em Ecotoxicology and Environmental Safety}, vol.~102, pp.~129--135, 2014.

\bibitem{9}
C.~Zhang, B.~Huang, J.~D. Piper, and R.~Luo, ``Biomonitoring of atmospheric
  particulate matter using magnetic properties of salix matsudana tree ring
  cores,'' {\em Science of The Total Environment}, vol.~393, no.~1,
  pp.~177--190, 2008.

\bibitem{10}
M.~Liao, S.~Kelley, and Y.~Yao, ``Generating energy and greenhouse gas
  inventory data of activated carbon production using machine learning and
  kinetic based process simulation,'' {\em ACS Sustainable Chemistry \&
  Engineering}, vol.~8, no.~2, pp.~1252--1261, 2020.

\bibitem{11}
P.~C. Austin, J.~Frank E~Harrell, and E.~W. Steyerberg, ``Predictive
  performance of machine and statistical learning methods: Impact of
  data-generating processes on external validity in the “large n, small p”
  setting,'' {\em Statistical Methods in Medical Research}, vol.~30, no.~6,
  pp.~1465--1483, 2021.
\newblock PMID: 33848231.

\bibitem{12}
P.~Karthikeyan, J.~Antony, and M.~Subagunasekar, ``Heavy metal pollutants and
  their spatial distribution in surface sediments from thondi coast, palk bay,
  south india,'' {\em Environmental Sciences Europe}, vol.~33, 12 2021.

\bibitem{13}
Y.~Fan, X.~Chen, Z.~Chen, X.~Zhou, X.~Lu, and J.~Liu, ``Pollution
  characteristics and source analysis of heavy metals in surface sediments of
  luoyuan bay, fujian,'' {\em Environmental Research}, vol.~203, p.~111911,
  2022.

\bibitem{14}
I.~Constantino, G.~Teodoro, A.~Moreira, F.~Paschoal, W.~Trindade, and
  M.~Bisinoti, ``Distribution of metals in the waters and sediments of rivers
  in central amazon region, brazil,'' {\em Journal of the Brazilian Chemical
  Society}, vol.~30, no.~9, pp.~1906--1915, 2019.

\bibitem{15}
R.~F. {dos Santos}, D.~Nagaoka, R.~B. Ramos, A.~B. Salaroli, S.~Taniguchi,
  R.~C.~L. Figueira, M.~C. Bícego, F.~J. Lobo, U.~Schattner, and M.~M. {de
  Mahiques}, ``Metal/ca ratios in pockmarks and adjacent sediments on the sw
  atlantic slope: Implications for redox potential and modern seepage,'' {\em
  Journal of Geochemical Exploration}, vol.~192, pp.~163--173, 2018.

\bibitem{17}
A.~G. Initiative, ``Australian and new zealand guidelines for fresh and marine
  water quality.''
\newblock [Accessed Dec. 8, 2021].

\bibitem{18}
R.~Altenburger, S.~Ait-Aissa, P.~Antczak, T.~Backhaus, D.~Barceló, T.-B.
  Seiler, F.~Brion, W.~Busch, K.~Chipman, M.~L. {de Alda}, G.~{de Aragão
  Umbuzeiro}, B.~I. Escher, F.~Falciani, M.~Faust, A.~Focks, K.~Hilscherova,
  J.~Hollender, H.~Hollert, F.~Jäger, A.~Jahnke, A.~Kortenkamp, M.~Krauss,
  G.~F. Lemkine, J.~Munthe, S.~Neumann, E.~L. Schymanski, M.~Scrimshaw,
  H.~Segner, J.~Slobodnik, F.~Smedes, S.~Kughathas, I.~Teodorovic, A.~J.
  Tindall, K.~E. Tollefsen, K.-H. Walz, T.~D. Williams, P.~J. {Van den Brink},
  J.~{van Gils}, B.~Vrana, X.~Zhang, and W.~Brack, ``Future water quality
  monitoring — adapting tools to deal with mixtures of pollutants in water
  resource management,'' {\em Science of The Total Environment}, vol.~512-513,
  pp.~540--551, 2015.

\end{thebibliography}

\end{document}